\documentclass[10pt,twocolumn,letterpaper]{article}

\usepackage[pagenumbers]{cvpr} %

\makeatletter
\@namedef{ver@everyshi.sty}{}
\makeatother
\usepackage{tikz}

\newcommand\RE[1]{\textcolor{red}{#1}}

\newcommand\MA[1]{\textcolor{magenta}{#1}}

\usepackage{colortbl}
\definecolor{sh_gray}{rgb}{0.84,0.84,0.84}
\definecolor{sh_gray2}{rgb}{1,0.89,0.75} %
\definecolor{color3}{rgb}{0.95,0.95,0.95}
\definecolor{color4}{rgb}{0.96,0.96,0.86}
\definecolor{color5}{rgb}{0.90,0.90,0.90}
\definecolor{tab_bkgd}{rgb}{0.88,0.88,0.88}

\usepackage{tikz}
\usetikzlibrary{backgrounds}
\usetikzlibrary{arrows,shapes}
\usetikzlibrary{tikzmark}
\usetikzlibrary{calc}
\usepackage{mathtools, nccmath}

\definecolor{attcolor}{rgb}{0. , 0.5 , 0.9}
\definecolor{convcolor}{rgb}{1 , 0.7 , 0}
\definecolor{mlpcolor}{rgb}{0.9 , 0.3 , 0.4}
\definecolor{othercolor}{rgb}{0.8 , 0.4 , 0.9}
\definecolor{attconvcolor}{rgb}{0. , 0.6 , 0.1}
\definecolor{wihtecolor}{rgb}{1 , 1. , 1.}
\definecolor{blackcolor}{rgb}{0, 0. , 0.}
\definecolor{rowColor}{rgb}{0.97, 0.97, 1}

\newcommand{\attshape}{\raisebox{0.5pt}{\tikz\fill[attcolor] (0,0) circle (.8ex);}}
\newcommand{\convshape}{\raisebox{0.5pt}{\tikz\fill[convcolor] (0,0) circle (.8ex);}}

\newcommand{\convattshape}{\raisebox{0.5pt}{\tikz\fill[attconvcolor](0,0) circle (.8ex);}}

\definecolor{newgreen}{rgb}{0, 0.6, 0.2}

\usepackage{amsmath}
\usepackage{amsthm}
\usepackage{bm}

\usepackage{dutchcal} %

\usepackage{algorithm}
\usepackage{algpseudocode}

\usepackage[hang,flushmargin]{footmisc} %

\usepackage{graphicx}
\usepackage{grffile} %
\usepackage{wrapfig} %
\usepackage{amssymb}

\usepackage{booktabs} %
\usepackage{array}
\usepackage{arydshln} %
\usepackage{multirow}
\usepackage{threeparttable} %
\usepackage{enumitem}
\usepackage{diagbox} %
\usepackage{makecell} %

\usepackage[pagebackref=true,breaklinks=true,letterpaper=true,colorlinks,bookmarks=false,citecolor=cyan,linkcolor=red]{hyperref}

\newcommand\bb[1]{\textbf{#1}}

\newcommand\x{$\times$}

\usepackage{pifont} %

\usepackage{soul} %

\begin{document}

\title{Why is the State of Neural Network Pruning \textit{so Confusing}? On the Fairness, Comparison Setup, and Trainability in Network Pruning}

\author{Huan Wang$^{\dag}$ \quad Can Qin \quad  Yue Bai \quad Yun Fu \\
Northeastern University, Boston, USA \\
}

\maketitle

{
\renewcommand{\thefootnote}{\fnsymbol{footnote}}
\footnotetext[2]{Corresponding author: \texttt{wang.huan@northeastern.edu}}
}

\begin{abstract}
The state of neural network pruning has been noticed to be unclear and even confusing for a while, largely due to ``a lack of standardized benchmarks and metrics''~\cite{blalock2020state}. To standardize benchmarks, first, we need to answer: \textbf{what kind of comparison setup is considered fair}? This basic yet crucial question has barely been clarified in the community, unfortunately. Meanwhile, we observe several papers have used (severely) sub-optimal hyper-parameters in pruning experiments, while the reason behind them is also elusive. These sub-optimal hyper-parameters further exacerbate the distorted benchmarks, rendering the state of neural network pruning even more obscure.

Two mysteries in pruning represent such a confusing status: the performance-boosting effect of a larger finetuning learning rate, and the no-value argument of inheriting pretrained weights in filter pruning. 

In this work, we attempt to explain the confusing state of network pruning by demystifying the two mysteries. Specifically, (1) we first clarify the fairness principle in pruning experiments and summarize the widely-used comparison setups; (2) then we unveil the two pruning mysteries and point out the central role of \textbf{network trainability}, which has not been well recognized so far; (3) finally, we conclude the paper and give some concrete suggestions regarding how to calibrate the pruning benchmarks in the future. Code:~\href{https://github.com/MingSun-Tse/Why-the-State-of-Pruning-So-Confusing}{https://github.com/mingsun-tse/why-the-state-of-pruning-so-confusing}.

\end{abstract}

\section{Introduction}
The past decade has witnessed the great success of deep learning, empowered by deep neural networks~\cite{LecBenHin15,schmidhuber2015deep}. The success comes at the cost of over-parameterization~\cite{alexnet,vgg,resnet,densenet,senet,vaswani2017attention,devlin2018bert,radford2019language,brown2020language,radford2021learning,rombach2022high}, causing prohibitive model footprint, slow inference/training speed, and rapid-growing energy consumption. Neural network pruning, an old model parameter reduction technique that used to be employed for improving generalization~\cite{baum1988size,chauvin1988back,Ree93}, now is mostly used for model size compression or/and speed acceleration~\cite{sze2017efficient,cheng2018recent,cheng2018model,gale2019state,deng2020model,hoefler2021sparsity}.

\begin{table}[t]
\centering
\caption{Top-1 accuracy (\%) benchmark of filter pruning with \textbf{ResNet50}~\cite{resnet} on \textbf{ImageNet}~\cite{imagenet}. Simply by using a better finetuning LR schedule, we manage to revive a \textit{6-year-ago} baseline filter pruning method, \textit{$L_1$-norm pruning}~\cite{li2017pruning}, making it \textit{match or beat} many filter pruning papers published in recent top-tier venues. Note, we achieve this simply by using the common step-decay LR schedule, 90-epoch finetuning, and standard data augmentation, \textit{no} any advanced training recipe (like cosine annealing LR) used. This paper studies the reasons and lessons behind this pretty confounding benchmark situation in filter pruning.}
\vspace{-3mm}
\resizebox{\linewidth}{!}{
\setlength{\tabcolsep}{1.3mm}
\begin{tabular}{lcc}
\toprule
Method & Pruned acc.~(\%) & Speedup \\
\midrule
SFP~\cite{he2018soft} $_\texttt{IJCAI'18}$ & 74.61 & 1.72$\times$  \\
DCP~\cite{zhuang2018discrimination} $_\texttt{NeurIPS'18}$ & 74.95 & 2.25$\times$ \\
GAL-0.5~\cite{lin2019towards} $_\texttt{CVPR'19}$ & 71.95 & 1.76$\times$ \\
Taylor-FO~\cite{molchanov2019importance} $_\texttt{CVPR'19}$ & 74.50 & 1.82$\times$ \\
CCP-AC~\cite{peng2019collaborative} $_\texttt{ICML'19}$ & 75.32 & 2.18$\times$ \\
ProvableFP~\cite{liebenwein2019provable} $_\texttt{ICLR'20}$ & 75.21 & 1.43$\times$  \\
HRank~\cite{lin2020hrank} $_\texttt{CVPR'20}$ & 74.98 & 1.78$\times$ \\
GReg-1~\cite{wang2021neural} $_\texttt{ICLR'21}$ & 75.16 & 2.31$\times$ \\
GReg-2~\cite{wang2021neural} $_\texttt{ICLR'21}$ & 75.36 & 2.31$\times$ \\
CC~\cite{li2021compact} $_\texttt{CVPR'21}$ & \bb{75.59} & 2.12$\times$ \\
$L_1$-norm~\cite{li2017pruning} $_\texttt{\RE{ICLR'17}}$ (\textbf{our reimpl.}) & 75.24 & \bb{2.31}$\times$ \\
\hdashline
GAL-1~\cite{lin2019towards} $_\texttt{CVPR'19}$ & 69.88 & 2.59$\times$ \\
Factorized~\cite{li2019compressing} $_\texttt{CVPR'19}$ & 74.55 & 2.33$\times$ \\
LFPC~\cite{he2020learning} $_\texttt{CVPR'20}$ & 74.46 & 2.55$\times$ \\
GReg-1~\cite{wang2021neural} $_\texttt{ICLR'21}$ & 74.85 & 2.56$\times$ \\
GReg-2~\cite{wang2021neural} $_\texttt{ICLR'21}$ & \bb{74.93} & 2.56$\times$ \\
CC~\cite{li2021compact} $_\texttt{CVPR'21}$ & 74.54 & \bb{2.68}$\times$ \\
$L_1$-norm~\cite{li2017pruning} $_\texttt{\RE{ICLR'17}}$ (\textbf{our reimpl.}) & 74.77 & 2.56$\times$ \\
\bottomrule
\end{tabular}}
\label{tab:teaser}
\vspace{-5mm}
\end{table}

The prevailing pipeline of pruning comprises three steps: 1) \bb{pretraining}: train a dense model; 2) \bb{pruning}: prune the dense model based on certain rules; 3) \bb{finetuning}: retrain the pruned model to regain performance. Most existing research focuses on the second step, seeking better criteria to remove unimportant weights so as to incur as less performance degradation as possible. This 3-step pipeline has been practiced for more than 30 years~\cite{mozer1988skeletonization,baum1988size,chauvin1988back,OBD} and is still extensively adopted in today's pruning methods~\cite{sze2017efficient,hoefler2021sparsity}. 

Despite the long history of network pruning, recent progress seems a little bit unclear. As we quote from the abstract of a recent paper~\cite{blalock2020state}, ``\textit{After aggregating results across 81 papers and pruning hundreds of models in controlled conditions, our clearest finding is that \ul{the community suffers from a lack of standardized benchmarks and metrics. This deficiency is substantial enough that it is hard to compare pruning techniques to one another or determine how much progress the field has made over the past three decades}}". We also sense this kind of confusion. Particularly, two mysteries in the area represent such confusion: 
\begin{itemize}
    \item \textbf{Mystery 1 (\texttt{M1}): The performance-boosting effect of a larger finetuning learning rate (LR)}. It \textit{was} broadly believed that finetuning the pruned model should use a \textit{small} LR, \eg, in the famous $L_1$-norm pruning~\cite{li2017pruning}, finetuning LR 0.001 is used for the ImageNet experiments. However, many other papers choose a larger LR, \eg, 0.01, which delivers significantly better performance than 0.001. For a pretty long time, pruning papers do not officially investigate this critical performance-boosting effect of a larger finetuning LR (although they may have \textit{already used} it in their experiments; see~Tab.~\ref{tab:summary_finetuning_lr_schedule}), until this paper~\cite{le2021network}. \cite{le2021network} formally studies how different finetuning LR schedules affect the final performance. They find random pruning and magnitude-pruning (the two most basic pruning methods) armed with a good finetuning LR schedule (CLR, or Cyclic Learning Rate Restarting) can counter-intuitively rival or even surpass other more sophisticated pruning algorithms. Unfortunately, they do not give explanations for this phenomenon except for calling upon comparing pruning algorithms in the same retraining configurations.
    \item \textbf{Mystery 2 (\texttt{M2}): The value of network pruning}. The value of network pruning seems unquestionable given the development history of over 30 years. However, two papers~\cite{crowley2018closer,liu2019rethinking} empirically find that training the pruned model from scratch can match pruning a pretrained model, thus radically challenging the necessity of pretraining a big model first in the conventional 3-step pruning pipeline.
\end{itemize}

Tab.~\ref{tab:teaser} shows a concrete example that \texttt{M1} makes the pruning benchmarks unclear. After using an improved finetuning LR schedule (see Tab.~\ref{tab:summary_finetuning_lr_schedule}), we make $L_1$-norm pruning~\cite{li2017pruning}, which is broadly considered the \textit{most basic baseline} approach, \textit{match or beat} many top-performing pruning methods published in top-tier conferences in the past several years. Such a situation really bewilders us, especially for those not in this area trying to borrow the most advanced pruning methods in their projects -- \textit{``Has the area really developed in the past several years?''} they might question.

This paper is meant to unveil these mysteries. Over this process, we hope to offer some helpful thoughts about why we have run into the current chaotic benchmark situation and how we can avoid it in the future. 

Specifically, when we try to unveil the mysteries, only to find there are various comparison setups enforced in the area, many of the conclusions actually \textit{hinge on which comparison setup is used}. To decide which comparison setup is more trustworthy, we have to be clear with the \textit{fairness principle} in pruning experiments first (\ie, what kind of comparison setup is considered fair?). After we sort out the fairness principle and comparison setups, we empirically examine the two mysteries and find \texttt{M2} reduces to \texttt{M1}. When examining \texttt{M1}, we find the role of \textit{network trainability} in network pruning, through which we can easily explain \texttt{M1}.

Therefore, our investigation path and the contributions in this paper can be summarized as follows. 
\begin{itemize}
    \item We first clarify the fairness principle in pruning experiments and summarize the outstanding comparison setups (Sec.~\ref{sec:fairness_comparison_setups}), which have been unclear in the literature for a long time.
    \item Then, we start to unveil \texttt{M1}~\cite{le2021network} and \texttt{M2}~\cite{liu2019rethinking}. As we shall show (Sec.~\ref{sec:value_of_pruning}), the conclusion of \texttt{M2} actually varies up to which comparisons setup is used: If a larger finetuning LR is allowed to be used, the no-value-of-pruning argument cannot hold; otherwise, it mostly holds. Thus, to unveil \texttt{M2}, we have to unveil \texttt{M1} first.
    \item Next, we focus on unveiling \texttt{M1}. We introduce the perspective of network trainability to diagnose pruning and clearly explain why the finetuning LR has a significant impact on the final performance (Sec.~\ref{sec:lr_effect}) -- to our best knowledge, we are the \textit{first} to clarify this mystery in the area.
    \item Finally, we summarize the major reasons that have led to the confusing benchmark situation of network pruning up to now, and give some concrete suggestions about how to avoid it in the future (Sec.~\ref{sec:conclusion}).
\end{itemize}

\section{Prerequisites} \label{sec:related_work}
\subsection{Taxonomy of Pruning and Related Work}
A pruning algorithm \textit{has and only has} five exclusive aspects to be specified: (1) Base model (when to prune): is pruning conducted on a pretrained model or a random model? (2) Sparsity granularity (what to prune): what is the smallest weight group in pruning? (3) Pruning ratio (how many to prune): how many weights are to be pruned? (4) Pruning criterion (by what to prune): what measure is used to select the important weights (\ie, those to be kept)~\vs~the unimportant weights (\ie, those to be pruned)? (5) Pruning schedule (how to schedule): How is the sparsity scheduled over the pruning process?

These five mutually orthogonal questions can shatter most (if not all) pruning papers in the literature. Researchers typically use (1), (2), and (4) to classify different pruning methods. We give the major backgrounds in these three axes.

\vspace{0.5em}
\noindent \bb{Pruning after training ~\vs~pruning at initialization}. Pruning has been mostly conducted on a \textit{pretrained} model over the past 30 years, which is thus called \textit{pruning after training} or \textit{post-training pruning}. This fashion has been the unquestioned norm until (at least) two papers in 2019, SNIP~\cite{lee2019snip} and LTH~\cite{frankle2019lottery}. They argue pruning can be conducted on \textit{randomly initialized} models and can achieve promising performance (allegedly matching the dense counterpart), too. This new fashion of pruning is called \textit{pruning at initialization} (PaI). Existing PaI approaches mainly include~\cite{lee2019snip,lee2020signal,wang2020picking,frankle2021pruning,ramanujan2020what} and the series of lottery ticket hypothesis~\cite{frankle2019lottery,frankle2020linear}. PaI is not very relevant to this work because the benchmarking chaos and the mysteries are mostly discussed in the PaT context, so here we would not discuss in length the specific PaI techniques. Interested readers may refer to~\cite{wang2021emerging} for a comprehensive summary.

\vspace{0.5em}
\noindent \bb{Sparsity structure: structured pruning~\vs~unstructured pruning}. If the smallest weight group in pruning is a single weight element, this kind of pruning is called unstructured (or fine-grained) pruning~\cite{han2015learning,han2015deep,frankle2019lottery}, because the resulting zero-weight (\ie, pruned-weight) locations are typically irregular (if no extra regularization is enforced). If the smallest weight group in pruning presents some structure, this kind of pruning is called structured (or structural/coarse-grained) pruning~\cite{wen2016learning,li2017pruning,he2017channel,mao2017exploring}. In the area, structured pruning typically narrowly refers to filter pruning or channel pruning if not explained otherwise. Structured pruning benefits more acceleration because the regular sparsity pattern is more hardware-friendly; 
 meanwhile, the regularity imposes more constraints on the network, so given the same sparsity level, structured pruning typically underperforms unstructured pruning. 
 
 Note, the definition of ``structured'' sparsity is severely hardware-dependent and thus can vary as the hardware condition changes. \Eg, the \textit{N:M sparsity}\footnote{https://developer.nvidia.com/blog/accelerating-inference-with-sparsity-using-ampere-and-tensorrt/} pioneered by NVIDIA Ampere architecture was considered as unstructured sparsity, but since NVIDIA has launched new library support to exploit such kind of sparsity for acceleration, now it can be considered as structured sparsity (called fine-grained structured sparsity~\cite{zhou2021learning,oh2022attentive}), too. 

This paper focuses on \textit{filter pruning} for now because the two aforementioned mysteries (the effect of finetuning LR and the value of network pruning) are mainly discussed in this context.

\vspace{0.5em}
\noindent \textbf{Pruning criterion: Importance-based~\vs~regularization based}. The former prunes weights based on some established importance criteria, such as magnitude (for unstructured pruning)~\cite{han2015learning,han2015deep} or $L_1$-norm (for filter pruning)~\cite{li2017pruning}, saliency based on 2nd-order gradients (\eg, Hessian or Fisher)~\cite{OBD,OBS,theis2018faster,wang2019eigendamage,singh2020woodfisher}. The latter adds a penalty term to the objective function, drives unimportant weights toward zero, then removes those with the smallest magnitude. Two notable points: (1) Even in a regularization-based pruning method, after the regularization process, the weights are still removed by a certain importance (typically magnitude). Namely, regularization-based pruning inherently embeds importance-based pruning. (2) The two paradigms are not exclusive; they can be employed simultaneously. \Eg,~\cite{DinDinHanTan18,wang2018structured,wang2021neural} select unimportant weights by magnitude while also employing the regularization to penalize weights.

Finally, for more comprehensive literature on network pruning, we refer interested readers to several surveys: an early one~\cite{Ree93}, some recent surveys of pruning alone~\cite{gale2019state,blalock2020state,hoefler2021sparsity} or pruning as a sub-topic under the general umbrella of model compression and acceleration~\cite{sze2017efficient,cheng2018recent,cheng2018model,deng2020model}.

\begin{figure*}[t]
\centering
\includegraphics[width=\linewidth]{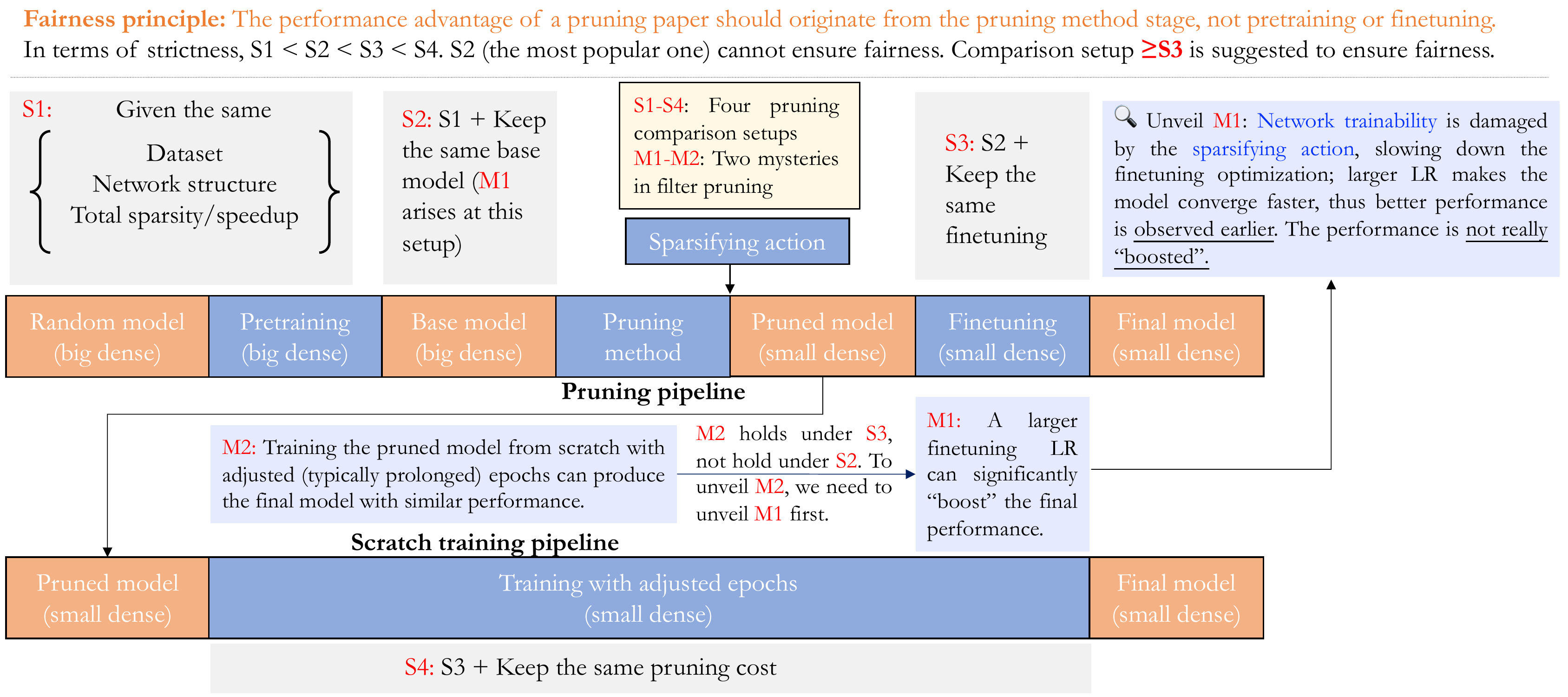} \\
\vspace{-1mm}
\caption{Overview of this paper. We are motivated by unveiling two mysteries (\texttt{M1}, \texttt{M2}) in filter pruning, which represent the confusing pruning benchmark situation. We first clarify the \textit{fairness principle} and summarize outstanding \textit{comparison setups} to lay down the discussion foundation (the notation ``\texttt{S1} $<$ \texttt{S2}'' means \texttt{S1} is \textit{less strict} than \texttt{S2}; others can be inferred likewise). Then we start to unveil \texttt{M1} and \texttt{M2}. \texttt{M2} will be shown to reduce to \texttt{M1}, actually. To unveil \texttt{M1}, we introduce \textit{network trainability} as an effective perspective to demystify \texttt{M1}. The unawareness of the role of network trainability in pruning has actually led to several sub-optimal hyper-parameter settings, which exacerbates the chaotic benchmark status. We finally give some concrete suggestions to calibrate the pruning benchmarks.}
\label{fig:overview}
\vspace{-5mm}
\end{figure*}

\vspace{0.5em}
\noindent \textbf{Most relevant papers}. The most relevant works to this paper are
\begin{itemize}
    \item \cite{blalock2020state}, which is the first to systematically report the frustrating state of network pruning benchmarks, identify some causes (such as the lack of standard benchmarks and metrics in pruning), and give concrete remedies. However, they do not go deeper and analyze why the standard benchmarks are hard to achieve. Our paper succeeds~\cite{blalock2020state} and will elaborate more on this aspect and point out the central role of trainability within.
    \item \cite{le2021network}, which officially reports the performance-boosting effect of a larger finetuning LR and calls upon comparing pruning algorithms under the same retraining configurations. \cite{le2021network} is actually motivated by~\cite{renda2020comparing}, which proposes \textit{LR rewinding}~\vs~the \textit{weight rewinding}~\cite{frankle2020linear} proposed for finding winning ticket on large-scale networks and datasets. \cite{renda2020comparing} virtually takes advantage of the performance-boosting effect of a larger finetuning LR, yet we do not know if they were aware at that point  that the performance boosting is not because of the magic rewound LR schedule but simply because of a larger finetuning LR (\ie, any appropriately larger LR would do, even not a rewound one), as later clarified by~\cite{le2021network}; so we tentatively consider \cite{le2021network} as the \textit{first} work to systematically report the performance-boosting effect of a larger finetuning LR.
    \item \cite{liu2019rethinking}, which brings forward the argument regarding the value of network filter pruning against scratch training. We will re-evaluate the major claim (scratch training can match filter pruning) of this paper under our more strictly controlled comparison setups.
\end{itemize}

\subsection{Terminology Clarification}
First, we make some critical concepts clear to lay down the common ground for discussion. Although they are pretty simple concepts, misinterpreting them will twist our discussions.
\begin{itemize}
    \item \textbf{\textit{Pruning pipeline~\vs~pruning method~\vs~sparsifying action}}. Some papers refer to \textit{pruning} as the whole pruning algorithm, \ie, the 2nd step in the pruning pipeline; while others may mean a pruning paper or the instant sparsifying action. We realize such a vague conception definition is one reason causing confusion, so we make them exact here. We use \textit{pruning pipeline} to mean all the three steps in a pruning paper. We can consider \textit{pruning pipeline} to be interchangeable with a \textit{pruning paper}. Then, we use \textit{pruning method = pruning algorithm} to mean the 2nd step of the pruning pipeline. Finally, the instant pruning action (\ie, zeroing out weights or physically taking away weights from a network) is referred to as~\textit{sparsifying action}. To summarize, in terms of concept scope, pruning paper = pruning pipeline = pruning $>$ pruning method = pruning algorithm $>$ sparsifying action\footnote{The notation ``pruning paper = pruning pipeline $>$ pruning method'' means, in this paper we consider \textit{pruning paper} interchangeable with \textit{pruning pipeline}, which includes \textit{pruning method} as one part.}.
    \item \textbf{\textit{Training from scratch}}. \textit{Training from scratch} = \textit{scratch training}, means to train a randomly initialized model to convergence. Scratch training of a \textit{pruned model} means, we already know the network architecture of the pruned model; the weights are randomly initialized; train this network from scratch using \textit{the same} training recipe as training the dense model. Notably, for filter pruning, when the architecture of the pruned model is known, the model should be implemented as a small-dense model, \textit{not a large-sparse model} (with structural masks).  The reason is, the widely-used parameter initialization schemes (\eg, He initialization~\cite{he2015delving}, the default initialization scheme for CONV and Linear layers in PyTorch~\cite{pytorch}) depend on the parameter shape. The large-sparse implementation is \textit{not} equivalent to (often underperforms) the small-dense implementation. For unstructured pruning, the standard implementation scheme is large-sparse weights (with unstructured masks)~\cite{liu2019rethinking,wang2021neural}.
    \item \textbf{\textit{Finetuning}}. After the sparsifying action action, the subsequent training process is called \textit{finetuning} or \textit{retraining}. We notice the community seems to have  different interpretations about these two terms. \Eg, in~\cite{le2021network}, finetuning is a \textit{sub-concept} of retraining, specifically meaning retraining with the last (smallest) learning rate of original training\footnote{\cite{le2021network} attributes this term usage to~\cite{han2015learning,li2017pruning,liu2019knowledge}. We double-checked these three papers and found they do not evidently have the inclination to mean finetuning as one particular type of retraining.}; while many more papers~\cite{wen2016learning,MolTyrKar17,liu2019rethinking,wang2019eigendamage,wang2021neural} consider finetuning \textit{the same as} retraining, meaning the 3rd step of the pruning pipeline. In this paper, we take the more common stance: considering finetuning and retraining interchangeable, and in the end of this paper, we will show the term ``finetuning'' should be deprecated in favor of ``retraining''.
    \item \textbf{\textit{Scratch-E, Scratch-B}}. These two terms are from~\cite{liu2019rethinking}, denoting two scratch training schemes. ``E'' is short for epochs, ``B'' short for (computation) budget. In practice, \cite{liu2019rethinking} uses FLOPs as an approximation for the computation budget. In Scratch-E, the point is to maintain the same \textit{total epochs} when comparing scratch training to pruning. In Scratch-B, the point is to maintain the same \textit{total FLOPs} when comparing scratch training to pruning. Here is a concrete example of Scratch-B: a dense model has FLOPs $F_1$; pretraining the dense model takes $K_1$ epochs; it is pruned by $L_1$-norm pruning, giving a pruned model with FLOPs $F_2$; the finetuning takes another $K_2$ epochs; then the scratch training should take $(K_1 F_1 + K_2 F_2) / F_2 = K_1(F_1/F_2) + K_2$ epochs. The ratio $F_1/F_2$ is typically called \textit{speedup} in network pruning.  
    \item \textbf{\textit{Value of network pruning}}. This term comes from~\cite{liu2019rethinking}. The conclusion of~\cite{liu2019rethinking} is that scratch training can match the performance of the 3-step pruning pipeline if Scratch-B is adopted, for filter pruning. Therefore, they argue there is no value for filter pruning algorithms that \textit{use predefined layerwise pruning ratios}. For the filter pruning algorithms that do not use  predefined layerwise pruning ratios, their role is to decide the favorable network architectures, akin to NAS~\cite{zoph2017neural,elsken2019neural}. As for unstructured pruning, \cite{li2017pruning} shows scratch training \textit{cannot} match pruning. Therefore, rigorously, the argument about the value of network pruning means the value of inheriting pretrained weights in filter pruning with predefined layerwise pruning ratios. We will use the short notion, \textit{value of network pruning}, without mentioning its much richer context.
\end{itemize}

\section{Fairness and Comparison Setups in Pruning} \label{sec:fairness_comparison_setups}
\subsection{\textit{Fairness Principle} in Network Pruning} \label{subsec:fairness}
This section is prepared for the next section, Sec.~\ref{subsec:comparison_setups}, where we will summarize the major comparison setups in pruning. As we shall see in the experiments, a pruning algorithm \texttt{A} can be better than \texttt{B} under one comparison setup, while worse than or on par with \texttt{B} under another comparison setup. To decide which comparison is more trustworthy, we have to evaluate which comparison setup is \textit{fairer}. Thence comes the necessity of a clear fairness principle in pruning.

\vspace{0.5em}
\noindent \textit{\textbf{Fairness Principle}. The performance advantage of a \ul{pruning paper} should be attributed to the \ul{pruning stage}, not finetuning or pretraining.} As aforementioned (Sec.~\ref{sec:related_work}), a pruning algorithm exclusively has five aspects~\cite{hoefler2021sparsity,wang2021emerging}: base model, sparsity granularity, pruning ratio, pruning criterion, and pruning schedule. Excluding the base model and sparsity granularity axes\footnote{Most pruning works do not consider the performance gain due to a better base model or sparsity granularity as a valid advantage over other methods, because a pruning method can be easily applied to different base models and sparsity granularities~(\eg, SSL~\cite{wen2016learning}) although the paper may only focus on one kind.}, therefore, \textbf{when we say a pruning method is better than another one, the performance advantage should be attributed to \textit{at least one of the three aspects}; namely, a better pruning ratio scheme, or/and a better pruning criterion, or/and a better pruning schedule}. Otherwise, it means the performance advantage comes from some outside factors other than the pruning algorithm itself -- in this case, attributing the performance credit to the pruning method would be an unjustified claim, potentially leading to an unfair comparison.

Per this fairness principle, clearly, we should keep the same base model and the finetuning process (\ie, the 1st and 3rd steps) in the pruning pipeline. Next, we elaborate on the outstanding comparison setups in network pruning and examine their fairness.

\subsection{Comparison Setups in Network Pruning} \label{subsec:comparison_setups}
In the literature, we find there are at least the following four groups of comparison setups, summarized in Tab.~\ref{tab:comparison_setups}. Note, in the following discussion, we consider different pruning algorithms (and scratch training) that remove or zero out unimportant weights \textit{only once}. Namely, we do not consider \textit{iterative pruning} for now, and we will discuss how the conclusions can carry over to iterative pruning.

\begin{table}[t]
\centering
\caption{Summary of popular comparison setups in pruning papers. It is helpful to review them along with the 3-step pruning pipeline: pretraining (output: base model) $\Rightarrow$ pruning (output: pruned model) $\Rightarrow$ finetuning (output: final small model). In terms of strictness, \texttt{S1} $<$ \texttt{S2}  $<$ \texttt{S3.1} $<$ \texttt{S3.2} $<$ \texttt{S4.1} $<$ \texttt{S4.2} (the notation ``\texttt{S1} $<$ \texttt{S2}'' means \texttt{S1} is \textit{less strict} than \texttt{S2}; others can be inferred likewise). Most existing pruning papers follow the \texttt{S2} comparison setup.}
\vspace{-3mm}
\resizebox{\linewidth}{!}{
\setlength{\tabcolsep}{0.5mm}
\begin{tabular}{ll}
\toprule
\textbf{No.} & \makecell[cc]{\textbf{Comparison setups}} \\
\midrule
\texttt{S1} & \makecell{Compare performance or performance drop on the same \\ dataset and network at the same compression or speedup rate} \\
\hline
\texttt{S2} & \makecell{+Same base model} \\
\hline
\texttt{S3.1} & \makecell{+Same base model\\+Same finetuning epochs} \\
\hdashline
\texttt{S3.2} & \makecell{+Same base model\\+Same finetuning LR schedule} \\
\hline
\texttt{S4.1} & \makecell{+Same base model\\+Same finetuning LR schedule\\+Same pruning epochs} \\
\hdashline
\texttt{S4.2} & \makecell{+Same base model\\+Same finetuning LR schedule\\+Same pruning LR schedule} \\
\hline
\texttt{SX-A} & \makecell{+Same epochs of ``pretraining + pruning + finetuning''} \\
\texttt{SX-B} & \makecell{+Same FLOPs of ``pretraining + pruning + finetuning''} \\
\bottomrule
\end{tabular}}
\label{tab:comparison_setups}
\vspace{-3mm}
\end{table}

\vspace{0.5em}
\noindent \textbf{Historical contexts of Tab.~\ref{tab:comparison_setups}.} In the following paragraphs, we briefly go through the historical context of the different comparison setups in~Tab.~\ref{tab:comparison_setups}.

\textbf{\MA{\texttt{S1}}}. Obviously, the \texttt{S1} setup is the most basic one, and also the earliest one. It simply compares performance regardless of many factors, such as training epochs and even the base model. This setup is adopted by early pruning papers, especially those using Caffe~\cite{jia2014caffe}, \eg,~\cite{han2015learning,han2015deep,he2017channel,wen2016learning}. Such a comparison setup does not even demand the same base model, which we may consider problematic today; yet back in those days, they had their own reasons -- before the deep learning (DL) community had mature DL platforms/tools/computation power as we have today, these papers usually \textit{trained their own base models}. Consequently, due to different implementations, their base models do not have the same (or even close) accuracy, \eg, the VGG16 base model reported by ThiNet~\cite{luo2017thinet} has top-5 accuracy 88.44\% while CP~\cite{he2017channel}, a concurrent work with ThiNet, reported 89.9\% (for those who are not familiar with these numbers, 1.5\% top-5 accuracy is a \textit{very significant} gap for ImageNet-1K classification).

As a remedy, to make the results comparable, many papers report the relative \textit{performance drop}, namely, base model accuracy minus final model accuracy. Such an idea is still broadly practiced at present~\cite{liu2019rethinking,wang2021neural}, \textit{esp.}~when comparing methods that are implemented under quite different conditions. 

\textbf{\MA{\texttt{S2}}}. Later, as the DL community develops, more DL platforms \eg, PyTorch~\cite{pytorch} and TensorFlow~\cite{abadi2016tensorflow} mature. There is usually a well-accepted model zoo (such as torchvision models\footnote{https://pytorch.org/vision/stable/models.html}) for others to use. As a result, more and more pruning papers adopt them as the base models, such as~\cite{li2017pruning,wang2021neural}, which has become the mainstream practice at present. Thus, the \texttt{S2} comparison setup arises.

\begin{table}[t]
\centering
\caption{Summary of \textit{finetuning} epochs and LR schedules of many filter pruning papers published in recent top-tier venues, with \textbf{ResNets}~\cite{resnet}. The default dataset is \textbf{ImageNet}~\cite{imagenet}; other datasets are explicitly pointed out.}
\vspace{-3mm}
\resizebox{\linewidth}{!}{
\setlength{\tabcolsep}{0.7mm}
\begin{tabular}{lcc}
\toprule
Method & \#Epochs & LR schedule \\
\midrule
SSL~\cite{wen2016learning}$_\texttt{NeurIPS'16}$ (CIFAR10) & -- & 0.01 \\
$L_1$-norm~\cite{li2017pruning}$_\texttt{ICLR'17}$ & 20 & 0.001, fixed \\
DCP~\cite{zhuang2018discrimination}$_\texttt{NeurIPS'18}$ & 60 & 0.01, step (36/48/54) \\
GAL-0.5/1~\cite{lin2019towards}$_\texttt{CVPR'19}$ & 30 & 0.01, step decay (10/20) \\
Taylor-FO~\cite{molchanov2019importance}$_\texttt{CVPR'19}$ & $\sim$25 & 0.01, step decay (10/20)  \\
Factorized~\cite{li2019compressing}$_\texttt{CVPR'19}$ & 90 & 0.01, step decay (30/60)  \\
CCP-AC~\cite{peng2019collaborative}$_\texttt{ICML'19}$ & 100 &  0.001, step decay (30/60/90) \\
HRank~\cite{lin2020hrank} $_\texttt{CVPR'20}$ & 30$\times\text{\#layers}$ & 0.01, step decay (10/20) \\
GReg-1/2~\cite{wang2021neural} $_\texttt{ICLR'21}$ & 90 & 0.01, step decay (30/60/75) \\
ResRep~\cite{ding2021resrep} $_\texttt{ICCV'21}$ & 180 & 0.01, cosine annealing \\
$L_1$-norm~\cite{li2017pruning} $_\texttt{ICLR'17}$ (\textbf{our reimpl.}) & 90 & 0.01, step decay (30/60/75) \\
\bottomrule
\end{tabular}}
\label{tab:summary_finetuning_lr_schedule}
\vspace{-5mm}
\end{table}

At this stage, few researchers have noticed the importance of finetuning. This makes sense since, in the pruning pipeline, only the pruning method part (\ie, the 2nd step) is regarded as the central one. The finetuning process is often so \textit{downplayed} that many papers do not even clearly report the hyper-parameters, as also noted by~\cite{blalock2020state}.

\textbf{\MA{\texttt{S3}}}. Later, in the endless pursuit of higher and higher performance, there is a clear trend that the finetuning epochs become longer and longer (see Tab.~\ref{tab:summary_finetuning_lr_schedule} for an incomplete summary). This in effect renders the comparison more and more unfair. Besides, the finetuning LR has been noticed to have a significant impact on the final performance, as formally studied by~\cite{le2021network} (although \cite{le2021network} is the first one to formally study this phenomenon, a larger finetuning LR has been employed by many papers even before). Because of these, the finetuning process must be taken into account to maintain fairness.

The most exact way to rule out the impact of finetuning is to use exactly the same finetuning process -- the same LR schedule (including the same epochs; we omit the hyper-parameters, like weight decay, momentum, \etc., and assume they are maintained the same), \ie, the \texttt{S3.2} in Tab.~\ref{tab:comparison_setups}. However, due to various objective or subjective reasons (\eg, prior papers may not release their finetuning details, making the follow-ups unable to reproduce the same finetuning), \texttt{S3.2} is often impractical.  Thence comes a weaker setup \texttt{S3.1}, which only keeps the same epochs of finetuning. It is allowed to use different finetuning LR schedules (\eg, different initial LR) -- \textit{this is where \texttt{M1}, the mystery of the finetuning LR effect, arises}.

Several papers~(such as~\cite{wang2021neural,wang2022trainability,zhang2021aligned,zhang2022learning}) have ablative analysis experiments on small-scale datasets (\eg, CIFAR10~\cite{cifar}) using the setup \texttt{S3.2}, while the main benchmark experiments (\eg, with ResNet-50 on ImageNet) using \texttt{S2}. The primary reason is that, following the setup \texttt{S3.2} means re-running the experiments for other comparison methods in the large-scale benchmarks, which is usually impractical (too costly) or even impossible (\eg, the comparison methods do not release usable code). 

\textbf{\MA{\texttt{S4}}}. The \texttt{S3.2} is still not the most strictly fair setup since it does not consider the cost (measured by the number of epochs) of the pruning method. For one-shot pruning (such as $L_1$-norm pruning~\cite{li2017pruning}), the cost of pruning is zero; while for a regularization-based method (such as GReg~\cite{wang2021neural}), it may take another few epochs for regularized training. Considering these cases, \texttt{S4.2} comes out: it builds upon \texttt{S3.2} and demands the same LR schedule for the pruning algorithm -- \textit{as far as we know, this is the most strict comparison setup}. In practice, again, for various reasons, we may not know the LR schedule of a pruning algorithm. Then, \texttt{S4.2} degrades to \texttt{S4.1}, which only demands the same epochs.

\textbf{\MA{\texttt{SX}}}. In setups \texttt{S2} to \texttt{S4.2}, when comparing pruning to scratch training in obtaining the same pruned (small) model, the scratch training employs the same training recipe of obtaining the base (big) model. \cite{liu2019rethinking} challenges this practice. They argue, the scratch training scheme spends less cost than pruning, so the comparison is unfair. As a remedy, they propose to take into account the cost of the \textit{pretraining} stage, which gives the \texttt{SX-A} and \texttt{SX-B} setups. About the \textit{cost}, one way to measure it is to use the number epochs (hence the \texttt{SX-A}); another is to consider the same computation and they approximate computation with FLOPs (hence the \texttt{SX-B}).  

It is hard to say if considering the cost of the pretraining stage is really necessary and practical. Advocates of the older practice may list reasons, \eg, pretrained models often exist already (like those pretrained on ImageNet~\cite{imagenet} and shared on HuggingFace\footnote{https://huggingface.co/}), so we do not need to consider the cost of pretraining. However, advocates of \texttt{SX} may argue that not all pretrained models are available; for many tasks, we still need to train the pretrained models first, so the cost of scratch training should be considered. 

We have no inclination here regarding which one is more correct. We make two points that we are fairly certain about: (1) In the pruning literature, most papers still follow the \textit{older} practice when reporting the scratch training results of the pruned model. (2) Given the recent rise of foundation models~\cite{bommasani2021opportunities} (\eg, Bert~\cite{devlin2018bert}, GPT-3~\cite{brown2020language}, CLIP~\cite{radford2021learning}, diffusion models~\cite{sohl2015deep,rombach2022high}), common researchers barely have the resources to train a model from scratch, so pruning would inevitably be conducted on the pretrained model, probably, for those big models.

\vspace{0.5em}
\noindent \bb{What comparison setup is mostly used now?} Unfortunately, \texttt{S2} is the most prevailing comparison setup at present~\cite{blalock2020state}. This setup ignores at least one important factor that, we now know~\cite{le2021network}, has a significant impact on the final performance: the finetuning LR schedule.

In the following sections, we start our empirical investigation of unveiling \texttt{M1} and \texttt{M2}. We study \texttt{M2} first and then \texttt{M1}, because the conclusion about \texttt{M2} actually depends on \texttt{M1}, as we are about to show.

\begin{table*}[t]
\centering
\caption{Top-1 accuracy (\%) comparison between $L_1$-norm pruning~\cite{li2017pruning} and training from scratch with \textbf{ResNet34} on \textbf{ImageNet100}. Each result is averaged by at least three random runs. The learning rate (LR) schedule of scratch training is: Initial LR 0.1, decayed at epoch 30/60/90/105 by multiplier 0.1, total: 120 epochs (top-1 accuracy of dense ResNet34: 84.56\%, FLOPs: 3.66G). \textit{``P30F90, 1e-1'' means the model is pruned at epoch 30 and finetuned for another 90 epochs with initial finetune LR 1e-1} (please refer to our supplementary material for the detailed LR schedule); the others can be inferred likewise. The \textbf{best} result within each comparison setup is highlighted \textbf{in bold}.}
\vspace{-2mm}
\resizebox{\linewidth}{!}{
\setlength{\tabcolsep}{3mm}
\begin{tabular}{lcccccc}
\toprule
Pruning ratio & 10\% & 30\% & 50\% & 70\% & 90\% & 95\% \\
FLOPs (G, speedup: $k\times$) & 3.30 (1.11$\times$) & 2.59 (1.41$\times$) & 1.90 (1.93$\times$) & 1.19 (3.09$\times$) & 0.48 (7.68$\times$) & 0.30 (12.06$\times$) \\
\midrule
Scratch training & 83.68$_{\pm0.38}$ & 83.31$_{\pm0.13}$ & 82.90$_{\pm0.16}$ & 82.45$_{\pm0.13}$ & 79.37$_{\pm0.76}$ & 76.67$_{\pm0.90}$ \\
\hline
\attshape~$L_1$-norm (P15F105, 1e-1) & 83.95$_{\pm0.17}$ & \bb{84.01}$_{\pm0.23}$ & \bb{83.87}$_{\pm0.44}$ & \bb{82.93}$_{\pm0.10}$ & 79.86$_{\pm0.11}$ & 77.41$_{\pm0.11}$ \\
\attshape~$L_1$-norm (P30F90, 1e-2) & 83.88$_{\pm0.07}$ & 84.00$_{\pm0.22}$ & 83.29$_{\pm0.14}$ & 82.61$_{\pm0.07}$ & \bb{80.41}$_{\pm0.32}$ & \bb{77.64}$_{\pm0.39}$ \\
\attshape~$L_1$-norm (P45F75, 1e-2) & 83.56$_{\pm0.03}$ & 83.95$_{\pm0.14}$ & 83.28$_{\pm0.08}$ & 82.47$_{\pm0.12}$ & 79.88$_{\pm0.10}$ & 76.17$_{\pm0.21}$ \\
\attshape~$L_1$-norm (P60F60, 1e-3) & 84.21$_{\pm0.07}$ & 83.87$_{\pm0.09}$ & 82.90$_{\pm0.10}$ & 81.24$_{\pm0.17}$ & 77.29$_{\pm0.05}$ & 70.53$_{\pm0.37}$ \\
\attshape~$L_1$-norm (P75F45, 1e-3) & \bb{84.24}$_{\pm0.04}$ & 83.47$_{\pm0.12}$ & 82.45$_{\pm0.14}$ & 80.81$_{\pm0.09}$ & 73.94$_{\pm0.24}$ & 64.98$_{\pm0.31}$  \\
\attshape~$L_1$-norm (P90F30, 1e-4) & 84.09$_{\pm0.07}$ & 82.47$_{\pm0.02}$ & 79.70$_{\pm0.00}$ & 74.87$_{\pm0.19}$ & 49.23$_{\pm0.21}$ & 29.89$_{\pm0.26}$ \\
\hdashline
\convshape~$L_1$-norm (P30F90, 1e-1) & \bb{85.27}$_{\pm0.13}$ & \bb{85.37}$_{\pm0.19}$ & \bb{85.48}$_{\pm0.18}$ & \bb{83.83}$_{\pm0.17}$ & \bb{81.56}$_{\pm0.29}$ & \bb{79.57}$_{\pm0.15}$ \\
\convshape~$L_1$-norm (P60F60, 1e-2) & 83.72$_{\pm0.14}$ & 83.88$_{\pm0.07}$ & 83.67$_{\pm0.11}$ & 82.96$_{\pm0.23}$ & 80.78$_{\pm0.23}$ & 77.81$_{\pm0.25}$ \\
\convshape~$L_1$-norm (P90F30, 1e-2) & 83.91$_{\pm0.08}$ & 84.02$_{\pm0.20}$ & 83.41$_{\pm0.15}$ & 82.91$_{\pm0.12}$ & 79.43$_{\pm0.07}$ & 75.20$_{\pm0.23}$ \\
\hdashline
\convattshape~$L_1$-norm (P30/$k$F90, 1e-1) & \bb{85.45}$_{\pm0.24}$ & \bb{85.06}$_{\pm0.24}$ & \bb{84.85}$_{\pm0.31}$ & \bb{83.64}$_{\pm0.09}$ & \bb{79.65}$_{\pm0.31}$ & \bb{75.79}$_{\pm0.28}$ \\
\convattshape~$L_1$-norm (P30/$k$F90, 1e-2) & 83.40$_{\pm0.04}$ & 82.69$_{\pm0.27}$ & 82.16$_{\pm0.03}$ & 79.97$_{\pm0.16}$ & 74.76$_{\pm0.24}$ & 70.61$_{\pm0.52}$ \\
\bottomrule
\end{tabular}}
\label{tab:imagenet100_l1_vs_scratch}
\begin{tablenotes}
    \item \small \attshape~Under comparison setup \texttt{S4.2} (same overall LR schedule), \convshape~Under comparison setup \texttt{SX-A} (same total epochs; finetuning LR increased), \convattshape~Under comparison setup \texttt{SX-B} (same total FLOPs).
\end{tablenotes}
\vspace{-5mm}
\end{table*}

\section{Reexamining the Value of Pruning} \label{sec:value_of_pruning}
The rethinking paper~\cite{liu2019rethinking} presents many valuable thoughts regarding the value of the 3-step pruning pipeline against scratch training. However, there are a few potential concerns in their experiments that may shake the validity of their conclusion. \textit{First}, they directly cite the results of a few pruning papers and compare the relative performance drop. Because of the stark differences between different DL platforms, such a comparison (\eg, comparing methods that use different base models) may not be convincing enough for rigorous analysis. \textit{Second}, when reproducing the $L_1$-norm pruning~\cite{li2017pruning}, they use fixed LR 0.001 and 20 epochs, following~\cite{li2017pruning}, for the finetuning stage, which is now known to be severely sub-optimal (see Tab.~\ref{tab:imagenet100_l1_vs_scratch}, a larger finetuning LR 0.01 can significantly boost performance). 

It is thereby of interest whether the no-value-of-pruning argument would change if the comparison is conducted under a strictly controlled condition and a better finetuning LR is employed. This section attempts to answer this question. Three comparison setups (\texttt{SX-A}, \texttt{SX-B}, and \texttt{S4.2}) are considered since they are the \textit{most strict} setups up to date.

\noindent \textbf{Pruning method}. We choose \textit{$L_1$-norm pruning}~\cite{li2017pruning} because it is the most representative pruning method and easy to control at a strict comparison setup. Specifically, $L_1$-norm pruning prunes the filters of a pretrained model with the smallest $L_1$-norms to a predefined pruning ratio. After pruning, the pruned model is finetuned for a few epochs to regain performance. Other pruning methods, such as regularization-based methods (\eg,~\cite{wen2016learning,liu2017learning,wang2021neural}), introduce many factors that are hard to control for rigorous analysis, so we do not adopt them here. We will discuss how the findings can generalize to those cases later.

\vspace{0.5em}
\noindent \textbf{Networks and datasets}. The network used for analysis is ResNet34~\cite{resnet}, following~\cite{li2017pruning}. For standard benchmarks (\eg, Tab.~\ref{tab:teaser}), we use ResNet50~\cite{resnet} because it is one of the most representative benchmark networks in filter pruning. The datasets are ImageNet100 and the full ImageNet~\cite{imagenet}. ImageNet100 is a randomly drawn 100-class subset of ImageNet. We use it for \textit{faster} analysis given our limited resource budget. The full ImageNet is used for benchmarks.

\vspace{0.5em}
\noindent \textbf{Implementation details of pruning}. For analysis, pruning is conducted on the 1st CONV layer (the 2nd CONVs are not pruned, following $L_1$-norm pruning~\cite{li2017pruning}) in all residual blocks of ResNet34. The first CONV and all FC layers are spared, also following the common practice~\cite{zhu2018prune,gale2019state,wang2021neural}. Uniform layerwise pruning ratio is employed (which usually under-performs a tuned non-uniform layerwise pruning ratio scheme; but since this paper does not target the best performance but explanation, we adopt it for easy analysis). We conduct pruning at a wide sparsity spectrum (10\% to 95\%) in the hopes of comprehensive coverage. 

\vspace{0.5em}
\noindent \textbf{One table to show them all}. The results are presented in Tab.~\ref{tab:imagenet100_l1_vs_scratch}. Before we present the analyses, we introduce a notion, \textit{pruning epoch}, which is defined as the epoch when the sparsifying action is physically enforced. \textit{E.g.}, if a model is trained for 30 epochs and then the sparsifying action is enforced, the pruning epoch is 30. We observe:

\textbf{(1)} For the \texttt{S4.2} setup (rows marked by~\attshape), we are not allowed to change the LR schedule. The only thing we can change is the pruning epoch. As seen, the best pruning epoch varies \textit{w.r.t.}~the sparsity level -- at a small pruning ratio, different pruning epochs give a similar performance; while as the pruning ratio arises, the performance becomes more sensitive to the pruning epoch, \eg, for pruning ratio 95\%, \texttt{P90F30, 1e-4} severely underperforms \texttt{P30F90, 1e-2}. Notably, a clear trend is, when the pruning ratio is large (70\% to 95\%), it is better to have a smaller pruning epoch.

Under this setup, only at pruning ratios of 30\%-70\%, pruning surpasses scratch training by a statistically significant gap. Therefore, we can only say pruning has a marginal advantage over scratch training here.  

\textbf{(2)} Then we look at the setup \texttt{SX-A} (rows marked by~\convshape). Under this setup, we are allowed to adjust the finetuning LR as long as the total epochs are kept the same. We increase the initial finetuning LR. As seen, it significantly improves the accuracies, \eg, (\texttt{P30F90, 1e-1}) improves the accuracy by nearly 2\% at pruning ratio 95\%, against (\texttt{P30F90, 1e-2}). This is the performance-boosting effect aforementioned~\cite{le2021network}. We also apply the larger LR trick to another two settings \texttt{P60F60} and \texttt{P90F30}. In all of them, we see a larger finetuning LR improves performance by an obvious margin.

Now, the gap between pruning and scratch training becomes much more significant. Pruning is more surely valuable under this setup.

\textbf{(3)} Next, we use the comparison setup \texttt{SX-B} (rows marked by~\convattshape), which maintains the total FLOPs. We apply this scheme to the best pruning setup \texttt{P30F90, 1e-1} in \texttt{S4.1} in the hopes of better performances. Since the dense model is trained for 30 epochs, to compensate for the FLOPs, the pruning epoch should be squeezed by the speedup ratio $k$. \textit{E.g.}, for pruning ratio 10\%, the speedup ratio is 1.11, then the pruning epoch should be adjusted to $30/k \approx 27$. 

As seen, the squeezing of the pruning epoch does close the gap between pruning and scratch training: At pruning ratios of 10\% to 70\%, pruning is still better; while for 90\% and 95\%, pruning only matches or underperforms scratch training -- this is a concrete example that we do \textit{not} have a once-for-all answer to questions like ``\textit{is pruning better than scratch training?}''

We also try a smaller finetuning LR in this setup, as shown in the row (\texttt{P30/$k$F90, 1e-2}). The LR effect also translates to this case -- a smaller finetuning LR degrades the performance.

\vspace{0.5em}
\noindent \textbf{Short summary}. As seen, the argument about the value of network pruning severely hinges on which comparison setup is employed and the pruning ratio. For the setup \texttt{SX-A}, where pruning outperforms scratch training obviously, the advantage comes from a better finetuning LR. Yet, we are not sure if such better LR schedules also exist for the scratch training; if so, scratch training can be further boosted, too -- as such, this kind of ``competition'' can be \textit{endless}. There are two kinds of attitudes toward this situation: \textit{(1)} Do not consider the performance improvement from a better finetuning LR as a fair/valid performance advantage as it is \textit{not} from the pruning algorithm. \textit{(2)} Still consider it as a valid performance advantage but will meet the ``endless competition'' challenge we just mentioned. The community now is mostly using \textit{(2)}. We suggest using \textit{(1)}, following our fairness definition clarified in Sec.~\ref{sec:fairness_comparison_setups}.

Despite many uncertainties, we are certain about one thing from Tab.~\ref{tab:imagenet100_l1_vs_scratch}: Whichever setup is favored, the finetuning LR holds a critical role to performance. Even for the comparisons setup \texttt{S4.2}, where the finetuning LR does not change, by changing the pruning epoch, implicitly, we change the finetuning LR, and it has been shown very pertinent to the final performance as well. In this sense, the two mysteries of pruning actually boil down to one (\texttt{M1}): \textit{Why does finetuning LR have such a great impact on the performance?}

This is the next question we would like to answer. LR, (arguably) as the most influential hyper-parameter in training neural networks, has a significant impact on performance -- this is definitely not surprising; what is really surprising might be, why the prior pruning works (\eg,~the original $L_1$-norm pruning~\cite{li2017pruning} adopts LR 0.001 in finetuning for their ImageNet experiments) did not realize that such a simple ``trick'' is so important to performance? This question is also worth our thinking.

\section{Trainability in Network Pruning} \label{sec:lr_effect}
\subsection{Background of Trainability}
Trainability, by its name, means the ability (easiness) of training a neural network, \ie, the optimization speed (note, speed is not equal to quality, so we may see a network with good trainability turns out to have a bad generalization ability eventually).

Notably, essentially, the role of a pruning method is to provide the initial weights for the later finetuning process, that is, pruning is essentially a kind of \textit{initialization}. In stark contrast to the broad awareness that initialization is very critical to neural network training~\cite{glorot2010understanding,sutskever2013importance,mishkin2015all,krahenbuhl2015data,he2015delving}, the initialization role of pruning has received negligible research attention, however. Trainability is also mostly studied for random initialization~\cite{saxe2014exact,xiao2018dynamical}.

A few recent works marry it with network pruning in some other similar forms like signal propagation~\cite{lee2020signal} and gradient flows~\cite{wang2020picking} (a good signal propagation or strong gradient flow usually suggests a good trainability). These works are inspiring, while they mostly stay in the domain of pruning at initialization (PaI). Few attempts before, to our best knowledge, tried to utilize the notion of trainability to examine pruning after training (PaT), at least, for the two mysteries we study here. This paper is meant to bridge this gap. The major difference between PaI and PaT is whether using a pretrained model as base. Such a context is essential to this paper since the above two mysteries are both brought forward in the context of PaT.

\begin{table}[t]
\centering
\caption{Top-1 accuracy (\%) comparison of different setups of $L_1$-norm pruning~\cite{li2017pruning} with \textbf{ResNet34} on \textbf{ImageNet100}. Pruning ratio: 95\%. TA: trainability accuracy (the metric used to measure trainability; see Eq.~(\ref{eq:trainability_acc})). This table shows, the performance gap between a smaller LR and a larger LR is not fundamental. It can be closed simply by training more epochs. The root cause that a smaller LR \textit{appears} to under-perform a larger LR is simply that the model trained by the smaller LR does \textit{not} fully converge.}
\vspace{-2mm}
\resizebox{\linewidth}{!}{
\setlength{\tabcolsep}{3mm}
\begin{tabular}{lcccccc}
\toprule
Finetuning setup & Top-1 acc. (\%) & TA (\%) \\
\midrule
P30F90, 1e-1 & 79.57$_{\pm0.15}$ & 88.00 \\
P30F90, 1e-2 & 77.64$_{\pm0.39}$ & 77.45 \\
P30F90, 1e-2 (+30 epochs) & 79.12$_{\pm0.19}$ & / \\
P30F90, 1e-2 (+60 epochs) & \bb{79.59}$_{\pm0.25}$ & / \\
\hdashline
P60F60, 1e-2 & \bb{77.81}$_{\pm0.25}$ & 87.39 \\
P60F60, 1e-3 & 70.53$_{\pm0.37}$ & 68.19 \\
P60F60, 1e-3 (+60 epochs) & 75.71$_{\pm0.09}$ & / \\
P60F60, 1e-3 (+120 epochs) & 77.17$_{\pm0.13}$ & / \\ 
P60F60, 1e-3 (+180 epochs) & 77.33$_{\pm0.09}$ & / \\ 
\hdashline
P90F30, 1e-2 & 75.20$_{\pm0.23}$ & 84.83 \\
P90F30, 1e-4 & 29.89$_{\pm0.26}$ & 37.93 \\
P90F30, 1e-4 (+60 epochs) & 60.69$_{\pm0.17}$ & / \\
P90F30, 1e-4 (+270 epochs) & 70.78$_{\pm0.16}$ & / \\
P90F30, 1e-4 (+1485 epochs) & \bb{78.18} & / \\
\bottomrule
\end{tabular}}
\label{tab:imagenet100_more_epochs}
\vspace{-5mm}
\end{table}

\vspace{0.5em}
\noindent \textbf{Trainability accuracy}. Literally, a bad trainability implies the training is hard and the training performance will arise slowly. Per this idea, there is a straightforward metric to measure trainability -- we introduce \textit{trainability accuracy}, the average of the first $N$ epochs,
\begin{equation}
T = \frac{1}{N} \sum_{i=1}^{N} \mathrm{Acc}_i.
\label{eq:trainability_acc}
\end{equation}
Since the optimization speed depends on the LR used, when we calculate trainability accuracy, we must ensure they are under the same LR schedule. In this paper, we choose $N$ as the number of the 1st LR stage, which characterizes the optimization speed in the early phase.

Next, we utilize trainability to explain the mysterious effect of the finetuning LR.

\subsection{Examining the Effect of Finetuning LR}
\noindent \textbf{Two facts as foundation}. We first lay down two facts as the common ground for the discussion of this section. We will show the mystery about the finetuning LR effect boils down to these two simple facts. 

\textit{First, pruning damages trainability}. This is an intuitively straightforward fact since pruning removes connections or neurons, which virtually makes the network harder to train. This fact holds for not only pruning a random network~\cite{lee2020signal}, but also for pruning a pretrained model here. Moreover, notably, more aggressive pruning leads to more damaged trainability. \textit{Second, a model of worse trainability will need more effective updates to reach convergence}. More effective updates mean two cases: If LR is not changed, more epochs are needed; if the number of epochs does not change, a larger LR is needed.  This is also easy to understand since trainability measures the easiness of optimization; a bad trainability implies harder optimization literally, hence the more effective updates. Such observation has been made by some sparse training papers, \eg, RigL~\cite{evci2020rigging} notes that ``\textit{sparse training methods benefit significantly from increased training steps}''.

When we observe that a larger LR improves the final test accuracy of the pruned model (\eg, Row \texttt{P30F90, 1e-1}~\vs~Row \texttt{P30F90, 1e-2} in Tab.~\ref{tab:imagenet100_l1_vs_scratch}), it is worthwhile to differentiate two subtle yet distinct possibilities:
\begin{itemize}
    \item A larger LR helps the pruned model reach a solution that the smaller LR \textit{cannot} reach, \ie, a larger LR help the model located into a better local minimum basin.
    \item The smaller LR can also help the model reach the solution as the larger LR does; just, the larger LR helps the model get there faster.
\end{itemize}
The former implies the performance-boosting effect of a larger LR is \textit{fundamental}; while the latter implies there is no fundamental gap between the two solutions; it is only an issue of optimization speed.

\begin{figure}[t]
\centering
\resizebox{0.8\linewidth}{!}{
\setlength{\tabcolsep}{1mm}
\begin{tabular}{ccccc}
\includegraphics{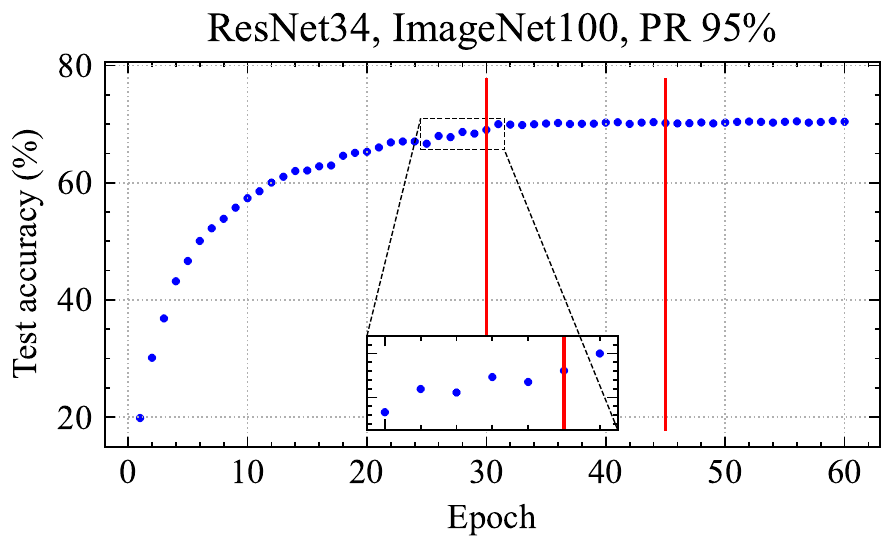} \\
(a) P60F60, 1e-3 \\
\includegraphics{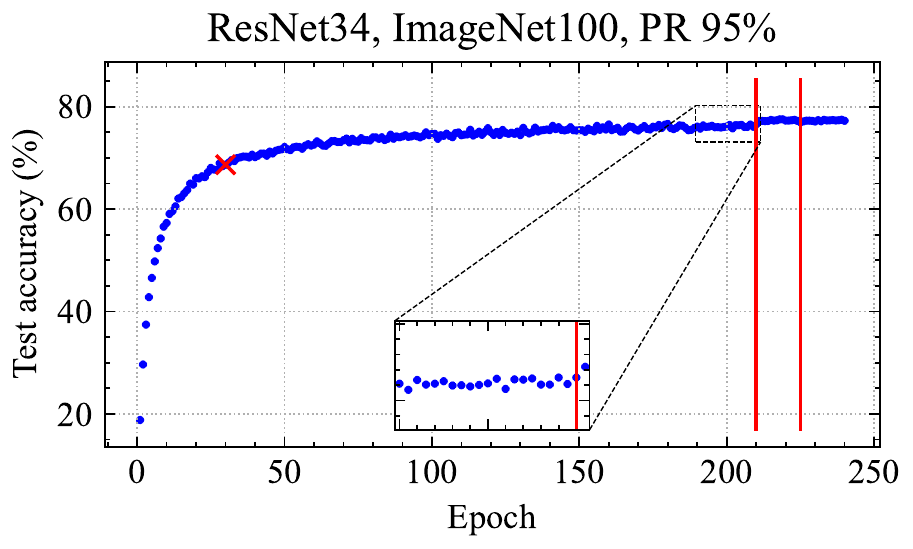} \\
(b) P60F60, 1e-3 (+180 epochs) \\
\end{tabular}}
\vspace{-3mm}
\caption{Test accuracy~\vs~epoch during finetuning of the setting \texttt{P60F60, 1e-3} at pruning ratio 95\% in Tab.~\ref{tab:imagenet100_more_epochs}. \RE{Red} vertical lines mark the epoch of decaying LR by 0.1. Particularly note before the 1st LR decay, the accuracy keeps arising in (a), implying the 1st LR decay may be too early -- this is confirmed in (b), where the red cross marker (\RE{$\times$}) indicates the time point of the 1st LR decay in (a). See more similar plots in our supplementary material.}
\label{fig:test_acc_vs_epoch}
\vspace{-5mm}
\end{figure}

Let's analyze a concrete example of \texttt{P60F60} in Tab.~\ref{tab:imagenet100_more_epochs}. For pruning ratio 95\% (we use this for example because at larger sparsity, the performance boosting effect is most pronounced), using 1e-2~\vs~1e-3 improves the test accuracy from 70.53 to 77.81, a very significant jump. This improvement also translates to the rows of \texttt{P30F90} and \texttt{P90F30}. 

However, in Tab.~\ref{tab:imagenet100_more_epochs}, we note the performance improvement coincides with trainability accuracy improvement. We were wondering if the performance improvement is actually due to a better trainability. 

Fig.~\ref{fig:test_acc_vs_epoch}(a) plots the test accuracy during the finetuning of \texttt{P60F60, 1e-3}. We notice before the 1st LR decay at epoch 30, the accuracy keeps arising even at epoch 30. This triggers a question: usually, we decay LR when the accuracy \textit{saturates}; now, when the accuracy is still steadily rising, \textit{is the LR decayed too early?} This question matters because if the LR decays too early, the model is \textit{forced} to stabilize due to the small step size and insufficient updates, not because it gets close to the local minimum, \ie, the model may \textit{not converge at all}.

To verify this, we extend the epochs of the LR 0.001 phase by 60/120/180 epochs. See the results in Tab.~\ref{tab:imagenet100_more_epochs} (note the rows \texttt{P60F60, 1e-3 (+60/120/180 epochs)}) and Fig.~\ref{fig:test_acc_vs_epoch}(b). Now, the model finetuned by LR 1e-3 can reach 77.33, very close to 77.81 reached by LR 1e-2. The test accuracy plot in Fig.~\ref{fig:test_acc_vs_epoch}(b) also confirms that the seeming underperformance of LR 1e-3 is due to insufficient epochs -- namely, \textbf{the advantage of a larger LR 0.01 is not some magic fundamental advantage, but a simple consequence of faster optimization.}
 
We also verify this on other cases (\texttt{P30F90} and \texttt{P90F30}) that the smaller LR ``underperforms'' the larger LR. The results are also presented in Tab.~\ref{tab:imagenet100_more_epochs}. In \textit{all} of these cases, given abundant epochs, the gap between the larger LR and the smaller LR can be closed. Especially for \texttt{P90F30}, the smaller LR 1e-4 can achieve a much better result than LR 1e-2 (78.18~\vs~75.20). This strongly demonstrates the smaller LR can also achieve what the larger LR can do.

To summarize, our results suggest \textbf{a larger LR does not really ``improve'' the performance. What really happens is, a larger LR accelerates the optimization process, making the higher performance \textit{observed earlier}}. In practice, when researchers tune different LR's, they usually keep the total epochs fixed (for the sake of fairness). Given the same total epochs, the pruned model using the smaller finetuning LR does not fully converge, making the performance \textit{appear} ``worse''.

\vspace{0.5em}
\noindent \textbf{Further remarks.} It is worthwhile to note that such an experimenting trap is \textit{so covert} if we are unaware of the damaged trainability issue in pruning. We may never realize that the epochs should be increased properly if a smaller finetuning LR is used. What's even trickier, we do not know how many more epochs is the so-called \textit{proper} -- Tab.~\ref{tab:imagenet100_more_epochs} is a living example. For some cases (\eg, \texttt{P30F90}), 60 more epochs is enough, while for others (\eg, \texttt{P60F60}, \texttt{P90F30}), 180 epochs or more is not enough to bridge the performance gap. Clearly, there is still much work to be done here toward a more rigorous understanding of the influence of damaged trainability on pruning.

\vspace{0.5em}
\noindent \textbf{Retrospective remarks and the lessons.} 
It is worthwhile to ponder why~\cite{li2017pruning} employed a \textit{seriously sub-optimal} finetuning LR scheme. This, we conceive, may originate from a long-standing \textit{misunderstanding} in the area of network pruning -- many have believed that because pruning is conducted onto a \textit{converged} model, the retraining of the pruned model needs not to be long and the LR should be \textit{small} to avoid destroying the knowledge the model has acquired, \eg, in~\cite{renda2020comparing}, the authors mentioned in their abstract ``\textit{The \ul{standard} retraining technique, fine-tuning, trains the unpruned weights from their final trained values using a \ul{small fixed learning rate}}'', implying that such misconception spreads so widely that it is taken for ``standard''.\footnote{Actually, the 3rd-step of the pruning pipeline is broadly referred to as \textit{finetuning} -- this term per se already implies the inclination of using a small LR. To rule out such conceptual bias, a more accurate way to phrase the 3rd step in the pruning pipeline may be \textit{retraining} the pruned model.} 

However, the results in Tab.~\ref{tab:imagenet100_l1_vs_scratch} suggest, such thought only holds for the cases of low pruning ratios. For a moderate or large pruning ratio, this thought \textit{hardly} holds. What was neglected is that the sparsifying action damages network trainability, slowing down the optimization. As a compensation, it is supposed to use a larger LR to accelerate the optimization, not a smaller LR; similarly, more epochs are needed to compensate for the slow optimization. However, the original $L_1$-norm pruning~\cite{li2017pruning} chose LR 0.001 and only 20 epochs for their ImageNet experiments, exactly the opposite of what is supposed. This, we conceive, is the reason that $L_1$-norm pruning has been underrated for a long time. Its real performance is actually pretty strong even compared with recent top-performing approaches (see Tab.~\ref{tab:teaser}).

Similarly, based on what we just learned about the truth of \texttt{M1}, if we examine the other filter pruning methods, \eg, GAL~\cite{lin2019towards} (see Tab.~\ref{tab:summary_finetuning_lr_schedule}), its reported results are probably also underrated, because it uses only 30 epochs for finetuning and the model may well not expose its full potential, as a result of the immature convergence. This implies a pretty disturbing concern -- for many filter pruning papers, we have to calibrate their results for fairness. Directly citing the numbers may (well) not show the real performance comparison.

\vspace{0.5em}
\noindent \textbf{How to make comparisons in a pruning paper?} As seen, different finetuning hyper-parameters can lead to very different conclusions, which do not reflect the true comparison of the pruning algorithm part. The ideal solution to this problem is to re-implement the existing pruning algorithms \textit{under the same configuration}. This counts on the community efforts (\eg,~ShrinkBench\footnote{https://github.com/JJGO/shrinkbench}\cite{blalock2020state} and Torch-Pruning\footnote{https://github.com/VainF/Torch-Pruning}~\cite{fang2023depgraph}) to standardize the benchmarks. Meanwhile, even with these efforts, we have to recognize that re-implementing all or most of the past pruning algorithms under the same condition is unrealistic, and sometimes unnecessary\footnote{\textit{Rigidly} demanding \textit{exact the same} condition in defining fairness may be of little meaning for scientific development. Think about the surge of deep learning -- compared to the past SVM era, deep learning nowadays definitely takes (way) more resources than SVM, \ie, unfair, while few has questioned the unprecedented value of deep learning as large language models like ChatGPT~\cite{chatgpt} surprise us on a daily basis.}. 

We suggest a more practical solution here, highlighting two rules of thumb: \textit{(1)} cross-validating different hyper-parameters; \textit{(2)} better performance still weighs more when fairness is hard to figure out. For a concrete example, given two pruning algorithms \texttt{A} (with finetuning setup $FT_A$) and \texttt{B} (with finetuning setup $FT_B$), we can conduct two more experiments:  \texttt{A} + $FT_B$ and \texttt{B} + $FT_A$; then compare \texttt{A} + $FT_A$~\vs~\texttt{B} + $FT_A$ and \texttt{A} + $FT_B$~\vs~\texttt{B} + $FT_B$. 
\begin{itemize}
    \item If one method consistently wins in both cases, we are more confident that the winner is really better due to its effective pruning design.
    \item If there is inconsistency between the two comparisons, \eg, \texttt{A} + $FT_A$ $>$ \texttt{B} + $FT_A$, while \texttt{A} + $FT_B$~$<$~\texttt{B} + $FT_B$, namely, different pruning methods favor different finetuning recipes. This means synergistic interaction exists between the pruning and finetuning stage and we cannot single out the finetuning when evaluating the pruning algorithms. In this case, we suggest that \textbf{the case delivering higher performance weighs more} as it advances the state-of-the-art.
\end{itemize}

\vspace{-0.5em}
\section{Conclusion and Discussion} \label{sec:conclusion}
\vspace{-0.5em}
This paper attempts to figure out the confounding benchmark situation in filter pruning. Two particular mysteries are explored: the performance-boosting effect of a larger finetuning LR, the no-value-of-pruning argument. We present a clear fairness principle and sort out four groups of popular comparison setups used by many pruning papers. Under a strictly controlled condition, we examine the two mysteries and find they both boil down to the issue of damaged network trainability. This issue was not well recognized by prior works, leading to (severely) sub-optimal hyper-parameters, which ultimately exacerbates the confounding benchmark situation in filter pruning now. We hope this paper helps the community towards a clearer understanding of pruning and more reliable benchmarks of it.

\vspace{0.5em}
\noindent \textbf{Takeaways and suggestions from this paper}:
\begin{itemize}
    \item \textit{Why is the state of neural network pruning so confusing?} Non-standard comparison setups (and its fundamental reason: unclear fairness principle) and the unawareness of the role of trainability are the two major reasons. The latter further leads to sub-optimal hyper-parameter settings, inherited by many follow-up papers, exacerbating the messy benchmark situation.
    \item As the area of network pruning develops, various comparison setups have appeared (see Tab.~\ref{tab:comparison_setups}). Each has its own historical context. Unfortunately, the most prevailing comparison setup now, setup \texttt{S2} in Tab.~\ref{tab:comparison_setups}, \textbf{cannot} ensure fairness. \textbf{We suggest using the setup \texttt{S3.2} or higher}, \ie, maintaining the same base model \textit{and the same finetuning process} -- Higher comparison setup means stricter experiment control, also means more resources and efforts; so there would be inevitably a \textit{trade-off} between how fair we want to be and how much we can invest in.
    \item Reporting all the finetuning details (\textit{esp.}~the LR schedule) is rather necessary and should be standardized.
    \item Cross-validating finetuning setups is a practical suggestion to decide the true winner between two pruning algorithms with different finetuning recipes.
    \item Filter pruning can beat scratch training or not, up to the specific comparison setup and pruning ratio in consideration. Given the recent rise of large foundation models, pruning may still follow the conventional 3-step pipeline.
    \item The observation that a larger finetuning LR ``improves'' pruning performance is largely a \textit{misinterpretation} -- the performance is not ``improved''; what really happens is that the good performance is observed earlier because the larger LR accelerates the optimization. The fundamental factor playing under the hood is the  network trainability damaged by the sparsifying action (or zeroing out) action in pruning.
    \item The damaged network trainability was not well recognized by prior pruning works, resulting in severely sub-optimal hyper-parameters, rendering the potential of a baseline method, $L_1$-norm pruning~\cite{li2017pruning}, \textit{underestimated} for a long time. This fact may spur us to \textit{re-evaluate} the  actual efficacy of (so) many sophisticated pruning methods against the simple $L_1$-norm pruning.     
    \item The term \textit{finetuning} (\ie, the 3rd step in the pruning pipeline) is not suitable. It has the incorrect conceptual bias of using a small LR -- we should firmly use \textit{retraining} instead.
\end{itemize}
{\small
\bibliographystyle{ieee_fullname}
\bibliography{egbib}
}

\appendix

\section{Training setting summary} \label{sec:train_setting_summary}
In the paper, we evaluate on two datasets ImageNet~\cite{imagenet} and ImageNet100. The latter is a small version (100 classes) of ImageNet for faster experimenting. We use PyTorch~\cite{pytorch} to implement all of our experiments. Therefore, we mainly refer to the official PyTorch ImageNet example\footnote{https://github.com/pytorch/examples/tree/main/imagenet} for hyper-parameters. The default LR schedule in PyTorch ImageNet example is: \texttt{0:0.1,30:0.01,60:0.001, total epochs:90}. This is adopted by $L_1$-norm pruning~\cite{li2017pruning} (and inherited by~\cite{liu2019rethinking}), so in their finetuning, they use the last-stage LR (\ie, 0.001) as finetuning LR and never decay it. 

\vspace{0.5em}
\noindent \textbf{How we decide the LR schedule for scratch training?} Per our empirical observations, decaying the LR from 1e-3 to 1e-4 can still improve the model by around 0.5-0.8\% top-1 accuracy. Namely, 1e-3 is \textit{not} where the model finally converges. Since we want to compare models \textit{at their best potential}, we add another 30 epochs and decay another two times to \textit{make sure the model fully utilizes its potential} -- this gives us the LR schedule for training a scratch model in Tab.~\RE{4}: \texttt{0:0.1,30:0.01,60:0.001,90:1e-4,105:1e-5, total epochs:120}. 

\vspace{0.5em}
\noindent \textbf{How we decide the LR schedule for finetuning?} We abide by the following rules to decide the finetuning LR schedule:
\begin{itemize}
    \item Rule 1: When comparing scratch training to pruning methods, \textbf{the final LR should be the same} -- as we mentioned, decaying LR from 1e-3 to 1e-4 on ImageNet can see another 0.5-0.8\% top-1 accuracy bump, thus it would be unfair to compare a pruning method whose model is finetuned to LR smaller than 1e-3 to a scratch-training model whose smallest LR is only 1e-3. The final LR for our ImageNet/ImageNet100 results, as mentioned above, is 1e-5.
    \item Rule 2: \textbf{Halving}. This means a kind of epoch splitting scheme: Given $N$ total epochs, split half of them (\ie, $N/2$ epochs) to the 1st LR, then split the \textit{half of the left epochs} (\ie, $N/4$) to the 2nd LR, \etc. To our best knowledge, this scheme is due to (no later than) the paper of ResNet~\cite{resnet} (see their CIFAR10 experiments).
    \item Rule 3: The epochs for each LR stage is \textbf{no more than 30}. This is mainly due to the design in the official PyTorch ImageNet example. We do not know why they chose 30 historically, but since this example is extensively followed as a baseline, we follow it too.
\end{itemize}
Based on these rules, given the total number of finetuning epochs, we can exactly derive the finetuning LR used in the paper. For example, in Tab.~\RE{4}, for \textit{\convshape P30F90, 1e-1}, the finetuning LR schedule is: \texttt{0:1e-1,30:1e-2,60:1e-3,75:1e-4,83:1e-5}; for \textit{\convshape P60F60, 1e-2}, the finetuning LR schedule is: \texttt{0:1e-2,30:1e-3,45:1e-4,53:1e-5}.

\vspace{0.5em}
\noindent \textbf{Layerwise pruning ratio for the experiments of ResNet50 on ImageNet}. For the results of ResNet50 on ImageNet we report in Tab.~\RE{1}, its finetuning LR schedule is: \texttt{0:1e-2,30:1e-3,60:1e-4,75:1e-5, total epochs:90}. As seen, this is never some heavily tuned magic LR schedule; nevertheless, we use it to finetune the pruned model by~$L_1$-norm pruning~\cite{li2017pruning}, only to find the final performance actually can \textit{beat/match many top-performing methods after ICLR'17}. The layerwise pruning ratios are borrowed from GReg~\cite{wang2021neural} (as they released their ratios; for many other papers, we only know the total speedup, not aware of the detailed layerwise pruning ratios) to keep a fair comparison with it -- speedup 2.31$\times$: [0, 0.60, 0.60, 0.60, 0.21, 0]; speedup 2.56$\times$: [0, 0.74, 0.74, 0.60, 0.21, 0].

\vspace{0.5em}
\noindent \textbf{Code references}. 
We mainly refer to the following code implementations in this work. They are all publicly available.
\begin{itemize}
    \item Official PyTorch ImageNet example\footnote{https://github.com/pytorch/examples/tree/master/imagenet};
    \item GReg-1/GReg-2~\cite{wang2021neural}\footnote{https://github.com/MingSun-Tse/Regularization-Pruning};
    \item Rethinking the value of network pruning~\cite{liu2019rethinking}\footnote{https://github.com/Eric-mingjie/rethinking-network-pruning/tree/master/imagenet/l1-norm-pruning}.
\end{itemize}

\section{Can the findings generalize to other pruning methods than $L_1$-norm pruning?} \label{sec:discussion_more_pruning_methods}
Pruning methods, according to their methodology, typically are classified into two groups, regularization-based (a.k.a.~penalty-based) and importance-based (a.k.a.~saliency-based), from a long time ago~\cite{Ree93}. 

Despite the different categorization, \textit{any} pruning method has a step to physically zero out the weights, \ie, the \textit{sparsifying action} step, per the terminology in this paper. Typically, this step is the magnitude pruning (or $L_1$-norm pruning when it comes to filter pruning). \textit{E.g.}, SSL~\cite{wen2016learning} and GReg~\cite{wang2021neural} are two regularization-based pruning methods, with different penalty terms proposed, yet both of them have a step to physically zero out unimportant weights by sorting their magnitude before finetuning. In other words, regularization-based methods, although they are classified into a different group from magnitude pruning (which is importance-based), they essentially include magnitude pruning as a part.

We have analyzed in the paper, the fundamental reason that incurs damaged trainability is the sparsifying action action in magnitude pruning. Therefore, \textit{any pruning method that employs magnitude pruning as a part is subject to the analyses of this paper} -- this means the discoveries of this paper are \textit{generic}. The attended broken trainability in these methods should also lead to \textit{similar}\footnote{This said, the severity of the trainability issue may vary up to specific pruning methods. \textit{E.g.}, we have observed the GReg method~\cite{wang2021neural} is less bothered by such damaged trainability due to their growing regularization design.} finetuning LR effect to the $L_1$-norm pruning case.

\section{Can the finetuning LR effect generalize to other LR schedules than the traditional step decay?} \label{sec:discussion_more_lr_schedules}
In the paper, we explore the finetuning LR effect (a large LR~\vs~a small LR,\eg, 0.01~\vs~0.001) to the final performance using the conventional \textit{step decay} LR schedule. It is of interest if the effect can translate to other more advanced LR schedules.

We consider \textit{Cosine Annealing LR schedule}~\cite{loshchilov2017sgdr} here, referring to the official PyTorch Cosine LR implementation\footnote{https://pytorch.org/docs/stable/generated/torch.optim.lr\_scheduler.CosineAnnealingLR.html}. When we switch from Step LR schedule to Cosine, the initial LR and minimum LR are kept the same (namely, the start point and end point of LR are the same; the only difference is the scheduling in between). The scratch model is trained for $200$ epochs, initial LR $0.1$, step decayed at epoch $100$ and $150$ by multiplier 0.1 (referring to the original ResNet CIFAR10 training recipe in the ResNet paper~\cite{resnet}). For finetuning, the initial LR is $0.01$ or $0.001$, the minimum LR $0.0001$, total epochs $120$.

\begin{table}[!h]
\centering
\caption{Effect of LR schedule of ResNet56 on CIFAR10. Baseline accuracy 93.78\%, Params: 0.85M, FLOPs: 0.25G.}
\vspace{-3mm}
\resizebox{\linewidth}{!}{
\setlength{\tabcolsep}{1mm}
\begin{tabular}{lcccccccccc}
\toprule
Pruning ratio & 0.3 & 0.5 & 0.7 & 0.9 \\
Sparsity/Speedup        & 31.14\%/1.45\x    & 49.82\%/1.99\x    & 70.57\%/3.59\x    & 90.39\%/11.41\x   \\
\midrule
Scratch (Step LR)                               & 93.16$_{\pm0.16}$ & 92.78$_{\pm0.23}$ & 92.11$_{\pm0.12}$ & \textbf{88.36}$_{\pm0.20}$ \\
Scratch (Cosine LR)                             & \textbf{93.84}$_{\pm0.06}$ & \textbf{93.20}$_{\pm0.31}$ & \textbf{92.15}$_{\pm0.21}$ & 88.17$_{\pm0.43}$ \\  
\hdashline
$L_1$~\cite{li2017pruning} (Step LR 0.001)      & 93.43$_{\pm0.06}$ & 93.12$_{\pm0.10}$ & 91.77$_{\pm0.11}$ & \textbf{87.57}$_{\pm0.09}$ \\
$L_1$~\cite{li2017pruning} (Step LR 0.01)       & \textbf{93.79}$_{\pm0.06}$ & \textbf{93.51}$_{\pm0.07}$ & \textbf{92.26}$_{\pm0.17}$ & 86.75$_{\pm0.31}$ \\
\hdashline
$L_1$~\cite{li2017pruning} (Cosine LR 0.001)    & 93.48$_{\pm0.04}$ & 93.11$_{\pm0.09}$ & 91.65$_{\pm0.11}$ & \textbf{87.17}$_{\pm0.14}$ \\ 
$L_1$~\cite{li2017pruning} (Cosine LR 0.01)     & \textbf{93.82}$_{\pm0.07}$ & \textbf{93.74}$_{\pm0.06}$ & \textbf{92.27}$_{\pm0.00}$ & 86.90$_{\pm0.20}$ \\
\bottomrule
\end{tabular}}
\label{tab:res56_cifar10_cosine_LR}
\vspace{-5mm}
\end{table}

\begin{table}[!h]
\centering
\caption{Effect of LR schedule of VGG19 on CIFAR100. Baseline accuracy: 74.02\%, Params: 20.08M, FLOPs: 0.80G.}
\vspace{-3mm}
\resizebox{\linewidth}{!}{
\setlength{\tabcolsep}{1mm}
\begin{tabular}{lcccccccccc}
\toprule
Pruning ratio & 0.3 & 0.5 & 0.7 & 0.9 \\
Sparsity/Speedup    & 19.24\%/1.23\x    & 51.01\%/1.97\x    & 74.87\%/3.60\x    & 90.98\%/8.84\x    \\
\midrule
Scratch (Step LR)   & 72.84$_{\pm0.25}$ & \textbf{71.88}$_{\pm0.14}$ & \textbf{70.79}$_{\pm0.08}$ & \textbf{66.52}$_{\pm0.37}$ \\
Scratch (Cosine LR) & \textbf{73.54}$_{\pm0.22}$ & 71.87$_{\pm0.09}$ & 70.10$_{\pm0.24}$ & 65.92$_{\pm0.10}$ \\
\hdashline
$L_1$~\cite{li2017pruning} (Step LR 0.001)  & 73.67$_{\pm0.05}$ & 72.04$_{\pm0.12}$ & 70.21$_{\pm0.02}$ & 64.72$_{\pm0.17}$ \\
$L_1$~\cite{li2017pruning} (Step LR 0.01)   & \textbf{74.01}$_{\pm0.18}$ & \textbf{73.01}$_{\pm0.22}$ & \textbf{71.49}$_{\pm0.14}$ & \textbf{66.05}$_{\pm0.04}$ \\
\hdashline
$L_1$~\cite{li2017pruning} (Cosine LR 0.001) & 73.69$_{\pm0.08}$ & 72.10$_{\pm0.08}$ & 69.96$_{\pm0.09}$ & 63.93$_{\pm0.15}$ \\
$L_1$~\cite{li2017pruning} (Cosine LR 0.01)  & \textbf{74.39}$_{\pm0.07}$ & \textbf{73.51}$_{\pm0.18}$ & \textbf{71.78}$_{\pm0.21}$ & \textbf{65.70}$_{\pm0.11}$ \\
\bottomrule
\end{tabular}}
\label{tab:vgg19_cifar100_cosine_LR}
\end{table}

The results of ResNet56 (on CIFAR10) and VGG19 (on CIFAR100) are presented in Tab.~\ref{tab:res56_cifar10_cosine_LR} and Tab.~\ref{tab:vgg19_cifar100_cosine_LR}. As seen, the advantage of initial LR 0.01 over 0.001 does not only appear with the Step LR schedule, but also appears with the Cosine LR schedule in most cases (\textit{esp.}~for VGG19). This implies the finetuning LR effect is \textit{generic}, not limited to one particular LR schedule, which further highlights the importance of the topic we have been studying in the paper.

\begin{table}[t]
\centering
\caption{Top-1 accuracy (\%) comparison of different setups of $L_1$-norm pruning~\cite{li2017pruning} with \textbf{ResNet34} on \textbf{ImageNet100}. Pruning ratio: 95\%. This table shows, the performance gap between a smaller LR and a larger LR is not fundamental. It can be closed (or squeezed) simply by training more epochs. The root cause that a smaller LR \textit{appears} to under-perform a larger LR is simply that the model trained by the smaller LR does \textit{not} fully converge.}
\vspace{-2mm}
\resizebox{\linewidth}{!}{
\setlength{\tabcolsep}{1mm}
\begin{tabular}{lcccccc}
\toprule
Finetuning setup & Top-1 acc. (\%) & Trainability acc. (\%) \\
\midrule
P90F30, 1e-2 & 75.20$_{\pm0.23}$ & 84.83 \\
P90F30, 1e-4 & 29.89$_{\pm0.26}$ & 37.93 \\
P90F30, 1e-4 (+60 epochs) & 60.69$_{\pm0.17}$ & / \\
P90F30, 1e-4 (+270 epochs) & 70.78$_{\pm0.16}$ & / \\
P90F30, 1e-4 (+1485 epochs) & \bb{78.18} & / \\
\bottomrule
\end{tabular}}
\label{tab:more_imagenet100_more_epochs}
\vspace{-3mm}
\end{table}

\section{Additional results} \label{sec:more_results}
\noindent \textbf{1e-4~vs.~1e-2 for \textit{P90F30}}.  In Tab.~\RE{5} of the paper, we mention the performance gap between a small LR and a large LR is not fundamental, but a simple consequence of convergence speed under different LRs. When we add more finetuning epochs, the performance gap can be closed fully or by a large part for \textit{P30F90} and \textit{P60F60}; while on \textit{P90F30}, the gap is still obvious even after we add 270 epochs. 

Here we add even more, 1485 epochs, so that the number of the 1st LR stage is now 1500 epochs, exactly 100 times of the 1st-LR-stage epochs (\ie, 15 epochs) when using 1e-2 as initial LR. As we see, now LR 1e-4 can achieve 78.18 top-1 accuracy, which is significantly better than 75.20 achieved by LR 1e-2. This is yet another strong piece of evidence to show that the seeming performance gap between a large LR and a small LR is \textit{never} a gap that cannot be bridged, further confirming our opinion in the paper.

\begin{figure}[t]
\centering
\resizebox{0.8\linewidth}{!}{
\setlength{\tabcolsep}{1mm}
\begin{tabular}{ccccc}
\includegraphics{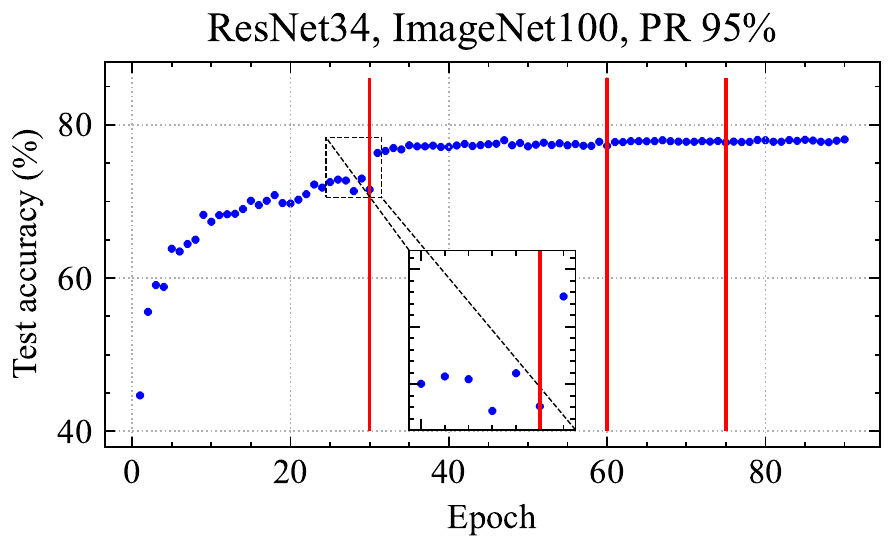} \\
(1.a) P30F90, 1e-2 \\
\includegraphics{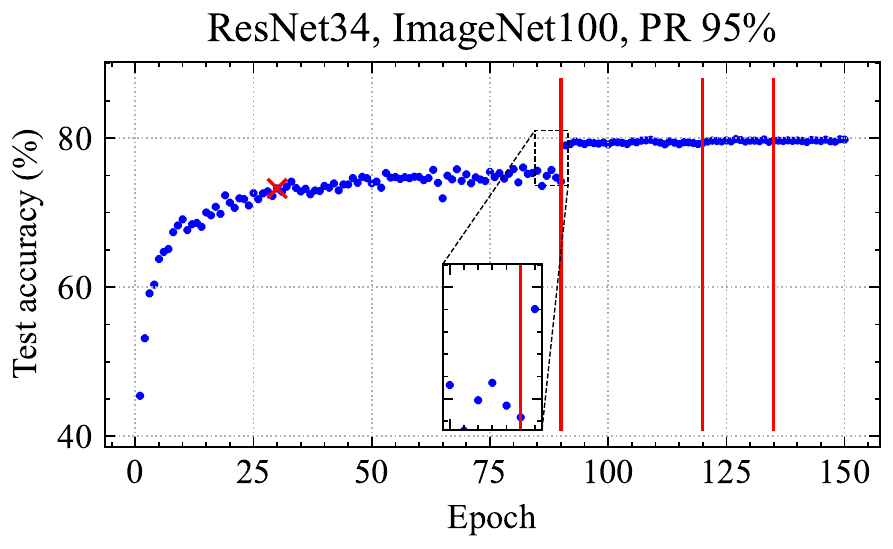} \\
(1.b) P30F90, 1e-2 (+60 epochs) \\
\includegraphics{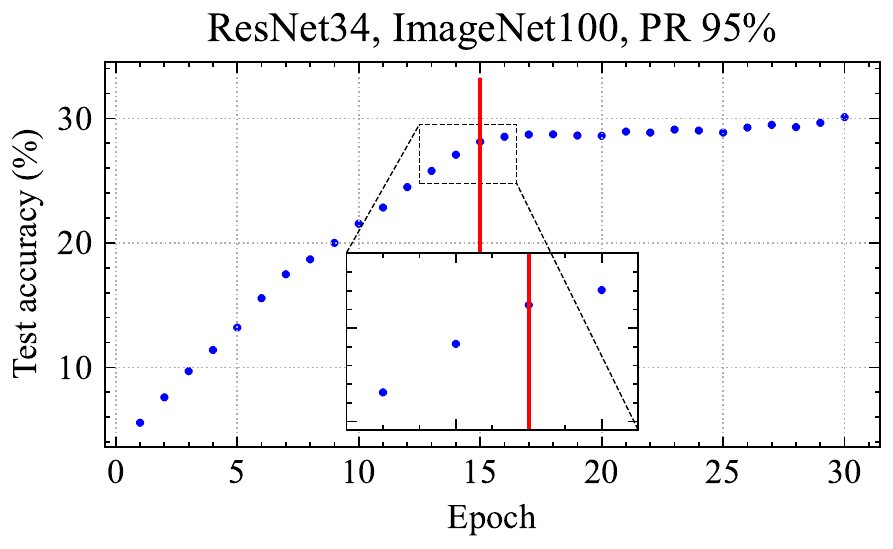} \\
(2.a) P90F30, 1e-4 \\
\includegraphics{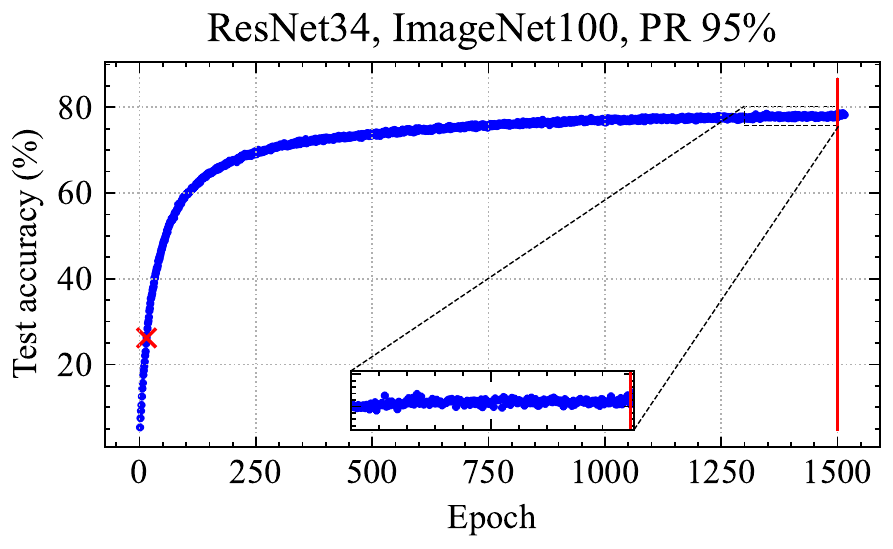} \\
(2.b) P90F30, 1e-4 (+1485 epochs) \\
\end{tabular}}
\vspace{-3mm}
\caption{Test accuracy~\vs~epoch during finetuning of the setting \textit{P30F90, 1e-2} and \textit{P90F30, 1e-4} at pruning ratio 95\% in Tab.~\RE{5}. \RE{Red} vertical lines mark the epoch of decaying LR by 0.1. Particularly note before the \textit{1st LR decay}, the accuracy keeps arising in (a), implying the 1st LR decay may be too early -- this is confirmed in (b), where the red cross marker (\RE{$\times$}) indicates the time point of the 1st LR decay in (a).}
\label{fig:more_test_acc_vs_epoch}
\end{figure}

\vspace{0.5em}
\noindent \textbf{More learning curves}. In Fig.~\RE{1} of the paper, we present the learning curves for \textit{P60F60, 1e-3} without and with more finetuning epochs, to show the underperformance of a small LR is actually due to insufficient training. Here we present more plots for \textit{P30F90, 1e-2} and \textit{P90F30, 1e-4} -- see Fig.~\ref{fig:more_test_acc_vs_epoch}.

In both cases (note the red crosses \RE{$\times$} in (1.b) and (2.b)), the performance can be boosted by adding more epochs to the 1st LR stage, especially for the case of \textit{P90F30, 1e-4}, where the 1st LR decay is actually \textit{way too early}. These plots further confirm our opinion in the paper -- the seeming underperformance of a small finetuning LR is not something magic, but a simple consequence of \textit{slow convergence} (caused by the broken trainability, \textit{esp.}~at large pruning ratios like 95\%).

\end{document}